\documentclass[conference]{IEEEtran}

\IEEEoverridecommandlockouts
\usepackage{cite}
\usepackage{amsmath,amssymb,amsfonts}
\usepackage{algorithmic}
\usepackage{graphicx}
\usepackage{textcomp}
\usepackage{hyperref}
\hypersetup{draft}
\usepackage{xcolor}
\def\BibTeX{{\rm B\kern-.05em{\sc i\kern-.025em b}\kern-.08em
    T\kern-.1667em\lower.7ex\hbox{E}\kern-.125emX}}

\usepackage{algorithm2e}
\usepackage{hadianeh}
\usepackage{macros}
\usepackage{nick_names}
\usepackage{results}
\newcommand{\subparagraph}{}
\usepackage{titlesec}
\usepackage[subtle,title]{savetrees}
\usepackage[square,sort&compress,comma,numbers]{natbib}

\begin{document}

\bstctlcite{paper:bstcontrol}

\author{
\IEEEauthorblockN{
\fontsize{10}{12}\selectfont{}Soroush Ghodrati \quad 
\fontsize{10}{12}\selectfont{}Hardik Sharma$^\dag$ \quad
\fontsize{10}{12}\selectfont{}Cliff Young$^\ddag$ \quad 
\fontsize{10}{12}\selectfont{}Nam Sung Kim$^\S$ \quad 
\fontsize{10}{12}\selectfont{}Hadi Esmaeilzadeh}
\\
\fontsize{10}{12}\selectfont{}\textbf{A}lternative \textbf{C}omputing \textbf{T}echnologies 
({\color[HTML]{0B6121}{\textbf{ACT}}}) Lab
\\
\vspace{0.05in}
\fontsize{10}{12}\selectfont{}University of California, San Diego \quad
\fontsize{10}{12}\selectfont{}$^\dag$Bigstream, Inc. \quad \fontsize{10}{12}\selectfont{}$^\ddag$Google Inc. \quad \fontsize{10}{12}\selectfont{}$^\S$University of Illinois Urbana-Champaign 
\\
\vspace{0.05in} 
\textcolor{blue}{\textsf{
\fontsize{10}{12}\selectfont \href{mailto:soghodra@eng.ucsd.edu}{soghodra@eng.ucsd.edu}\quad \fontsize{10}{12}\selectfont \href{mailto:hardik@bigstream.co}{hardik@bigstream.co} \quad \fontsize{10}{12}\selectfont \href{mailto:cliffy@google.com}{cliffy@google.com} \quad
\fontsize{10}{12}\selectfont \href{mailto:nskim@illinois.edu}{nskim@illinois.edu} \quad
\fontsize{10}{12}\selectfont \href{mailto:hadi@eng.ucsd.edu}{hadi@eng.ucsd.edu}
}}
}

\title{\huge Bit-Parallel Vector Composability for Neural Acceleration}

\date{}
\maketitle



\begin{abstract}
Conventional neural accelerators rely on isolated self-sufficient functional units that perform an atomic operation while communicating the results through an operand delivery-aggregation logic.
Each single unit processes all the bits of their operands atomically and produce all the bits of the results in isolation.
This paper explores a different design style, where each unit is only responsible for a slice of the bit-level operations to interleave and combine the benefits of bit-level parallelism with the abundant data-level parallelism in deep neural networks.
A dynamic collection of these units cooperate at runtime to generate bits of the results, collectively.
Such cooperation requires extracting new grouping between the bits, which is only possible if the operands and operations are vectorizable.
The abundance of Data-Level Parallelism and mostly repeated execution patterns, provides a unique opportunity to define and leverage this new dimension of Bit-Parallel Vector Composability.
This design intersperses bit parallelism within data-level parallelism and dynamically interweaves the two together.
As such, the building block of our neural accelerator is a Composable Vector Unit that is a collection of Narrower-Bitwidth Vector Engines, which are dynamically composed or decomposed at the bit granularity.
%
Using six diverse CNN and LSTM deep networks,  we evaluate this design style across four design points: with and without algorithmic bitwidth heterogeneity and with and without availability of a high-bandwidth off-chip memory.
%
Across these four design points, Bit-Parallel Vector Composability brings (\Speedupmin to \Speedupmax) speedup and (\Energymin to \Energymax) energy reduction.
%
%
We also comprehensively compare our design style to the Nvidia's RTX~2080~TI GPU, which also supports INT-4 execution.
The benefits range between \perfWattminRTX and \perfWattmaxRTX improvement in Performance-per-Watt.
%

%


\end{abstract}

\begin{IEEEkeywords}
Acceleration, neural networks, bit-flexibility
\end{IEEEkeywords}

\section{Introduction}
\label{sec:intro}

The growing body of neural accelerators~\cite{eyeriss:isca:2016,cnvlutin:isca:2016,stripes:micro:2016,eie:isca:2016, isaac:isca:2016, prime:isca:2016, scnn:isca:2017, tpu:isca:2017, bitfusion:isca18, unpu:isscc:2018, brainwave:isac:2018} exploit various forms of Data-Level Parallelism (DLP) that are abundant in Deep Neural Networks.
For instance, Google's TPU~\cite{tpu:isca:2017} extracts data-level parallelism across the lanes of its systolic array, Microsoft's Brainwave~\cite{brainwave:isac:2018} builds a dataflow architecture from vectorized units, and GPUs have been long designed for Single-Instruction Multiple-Data (SIMD) execution model.
Nonetheless, these various organization still rely on isolated self-sufficient units that process all the bits of input operands and generate all the bits of the results.
These values, packed as atomic words, are then communicated through an operand delivery-aggregation interconnect for further computation.
This paper sets out to explore a different design where each unit in vectorized engines is only responsible for processing a bit-slice.
This design offers an opening to explore the interleaving of Bit-Level Parallelism with DLP for neural acceleration.
Such an interleaving opens a new tradeoff space  where the complexity of Narrow-Bitwidth Functional Units needs to be balanced with respect to the overhead of bit-level aggregation as well as the width of vectorization.
\begin{figure}
	\centering
	\includegraphics[width=0.4\linewidth]{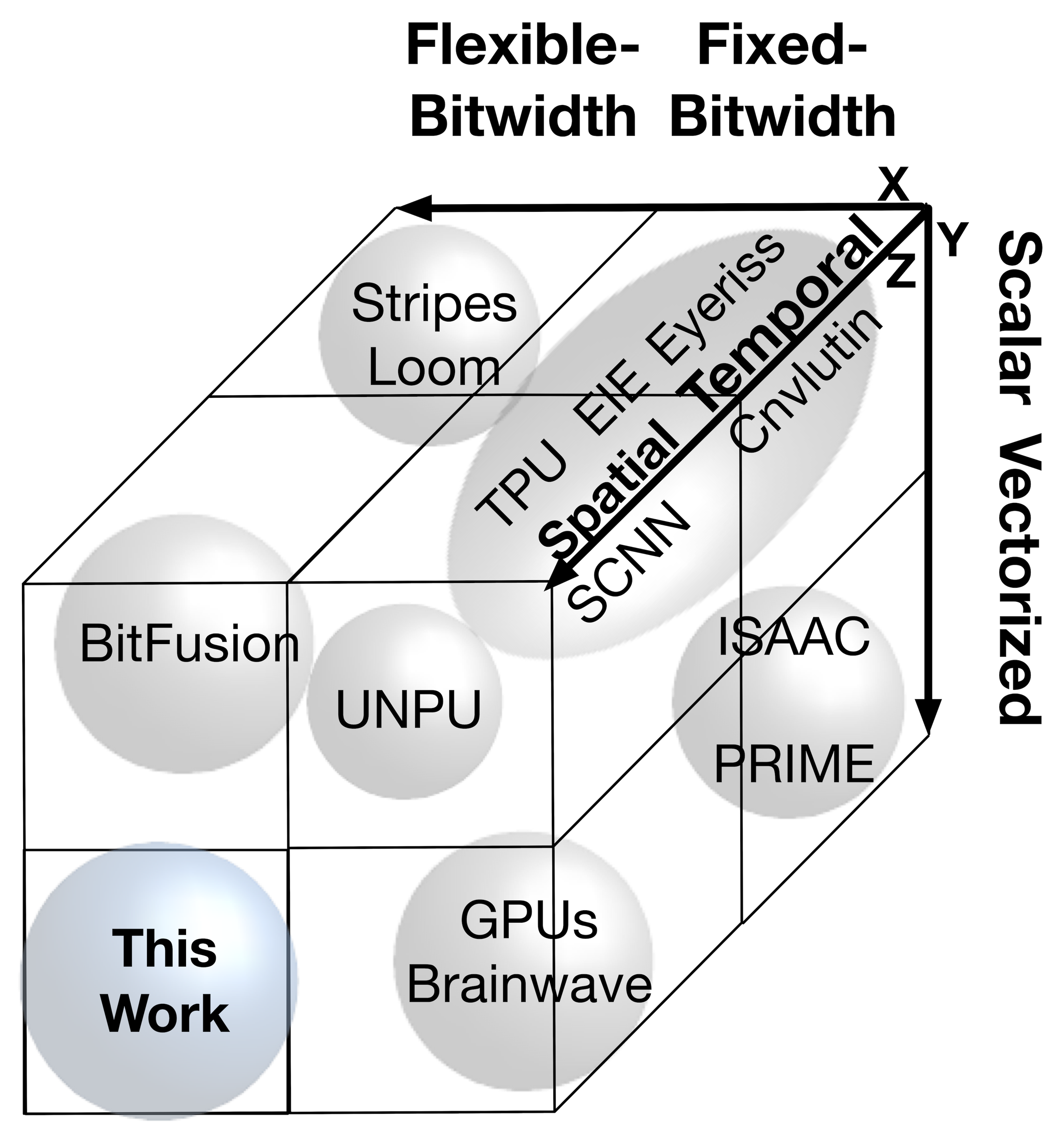} 
	\vspace{-1ex}
	\caption{The landscape of DNN accelerators and how this work fits in the picture.}
	\label{fig:intro}
 	\vspace{-4ex}
\end{figure}

Additionally, the proposed approach creates a new opportunity for exploring bit-flexibility in the context of vectorized execution.
As Figure~\ref{fig:intro} illustrates, the landscape of neural accelerators can be depicted in the three dimensions of Type of Functional Units (Scalar~\cite{eyeriss:isca:2016, eie:isca:2016, cnvlutin:isca:2016, tpu:isca:2017, scnn:isca:2017} v.s. Vectorized~\cite{isaac:isca:2016, prime:isca:2016, brainwave:isac:2018}), Bit 
Flexibility\footnote{Bit flexibility can be defined as the capability of a functional unit in reconfiguring itself \emph{at runtime} to support different bitwidths.} (Fixed Bitwidth v.s. Flexible Bitwidth), and Composability (Temporal~\cite{stripes:micro:2016, loom:arxiv:2017, unpu:isscc:2018} vs. Spatial~\cite{bitfusion:isca18}).
This work aims to fill the vacancy for Vectorized Bit-Flexible Spatial Composability.
In fact, the interleaving of the bit-level parallelism with DLP creates the new opportunity to explore bit-level composability where a group of bit-parallel vector units \emph{dynamically} form collaborative groups to carry out SIMD operations.
Recent inspiring research investigates bit-level composability for a scalar operation (a single MAC unit) both in spatial and temporal mode.
However, the space where composability (bit-level parallelism) meets vectorization (data-level parallelism) is not explored.
Our design style opens this space by breaking up the vector units to narrower bitwidth engines.
Since the building block is a narrow bitwidth vector engine, another opportunity arises to support execution well below eight bits.
That is leveraging the algorithmic insight that heterogenous assignment of bitwidths below eight to DNN layers can reduce their computational complexity while preserving accuracy~\cite{qnn:arxiv:2016, wrpn, choi2018pact, codex, releq}.
Fixed bitwidth designs cannot tap into this level of efficiency where each unit is only processing the fewest number of bits that can preserve the accuracy.

We evaluate the proposed concepts with and without availability of algorithmic bitwidth heterogeneity.
Experimentation with six real-world DNNs shows that bit-parallel vector composibility provides $40\%$ speedup and energy reduction compared to design with the same architecture (systolic) without support for the proposed composability.
When bit flexibility is applicable because of the heterogenous bitwidths in the DNN layers, our design provides $50\%$ speedup and $10\%$ energy reduction compared to BitFusion~\cite{bitfusion:isca18}, the state-of-the-art architecture that supports scalar bit-level composability in systolic designs\footnote{We do not compare to BitFusion when algorithmic bitwidth heterogeneity is not available, BitFusion would underperform compared to a non-composable design.}.
%

When a high bandwidth off-chip memory is utilized, the baseline design only enjoys $10\%$ speedup and $30\%$ energy reduction, respectively.
%
However, bit-parallel vector-composability better utilizes the boosted bandwidth and provides $2.1\times$ speedup and $2.3\times$ energy reduction.
With algorithmic bitwidth heterogeneity our design style provides $2.4\times$ speedup and $20\%$ energy reduction, compared to BitFusion, while it also utilizes the same high bandwidth memory
%
Finally, we compare different permutations of our design style with respect to homogenous/heterogenous bitwidth and off-chip memory bandwidth to the Nvidia's RTX~2080~TI GPU which also supports INT-4 execution.
%
Benefits range between \perfWattminRTX and \perfWattmaxRTX higher Performance-per-Watt for  four possible design points.
\section{Bit-Parallel Vector Composability}
\label{sec:overview}
%
This paper builds upon the fundamental property of \textit{vector dot-product operation} -- the most common operation in DNNs -- that vector dot-product with wide-bitwidth data types can be decomposed and reformulated as a summation of several dot-products with narrow-bitwidth data types.
The element-wise multiplication in vector dot-product can be performed independently, exposing \emph{data-level parallelism}.
This work explores another degree of parallelism -- \emph{bit-level parallelism} (BLP) -- wherein individual multiplications can be broken down at \emph{bit-level} and written as a summation of narrower bitwidth multiplications.
Leveraging this insight, this work studies the interleaving of both data-level parallelism in vector dot-product with bit-level parallelism in individual multiplications to introduce the notion of \emph{bit-parallel vector composability}.
%
%
%
This design style can also be exploited to support \textit{runtime-flexible} bitwidths on the underlying hardware.
The rest of this section details the mathematical formulation for bit-parallel vector composability. 

\begin{figure}[!b]
\vspace{-1ex}
	\centering
	\includegraphics[width=1\linewidth]{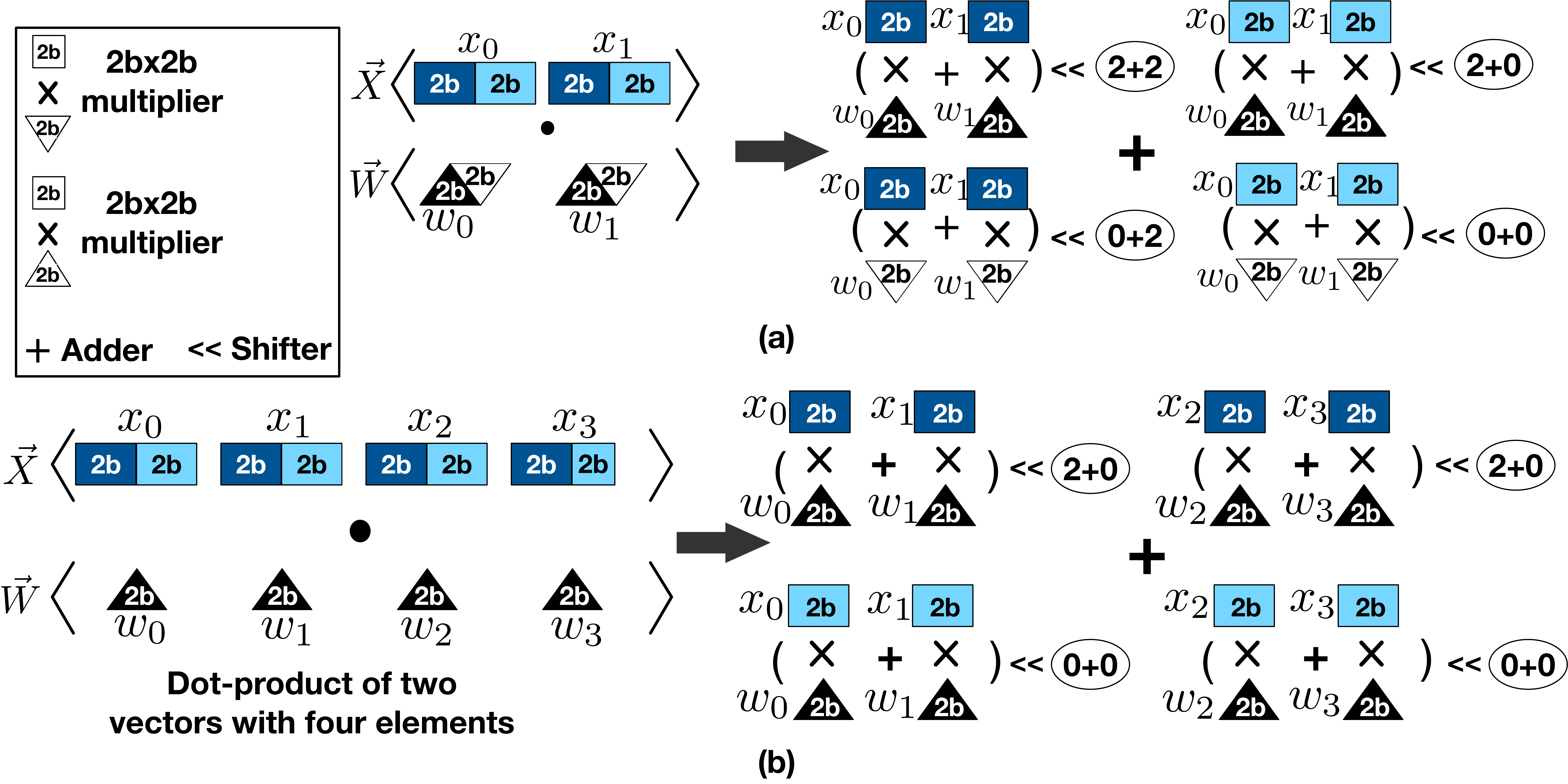} 
	\vspace{-2.5ex}
	\caption{(a) Fixed-bitwidth bit-parallel vector composability with 2-bit slicing and (b) Bit-Flexible vector composability for 4-b inputs and 2-b weights; 2x improvement in performance compared to fixed 4-bit dot-product.}
	\label{fig:compute_model}
\end{figure}

%
\niparagraph{Fixed-Bitwidth bit-parallel vector composability.}
Digital values can be expressed as the summation of individual bits multiplied by powers of two.
Hence, a vector dot-product operation between two vectors, $\vec{X}$, $\vec{W}$ can be expressed as follows:
\begin{equation}
	\resizebox{1\columnwidth}{!}{
	$\vec{X} \bullet \vec{W} = 
	\sum_i (x_i \times w_i) = \sum_i ((\sum_{j=0}^{bw_x-1}2^{j} \times x_i[j]) \times (\sum_{k=0}^{bw_w-1}2^{k} \times w_i[k]))$}
\label{eq:dot-product1}
\end{equation}
Variables $bw_x$ and $bw_w$ represent the bitwidths of the elements in $\vec{X}$ and $\vec{W}$, respectively.
Expanding the bitwise multiplications between the elements of the two vectors yields:
%
\begin{equation}
	\resizebox{0.75\columnwidth}{!}{
	$\vec{X} \bullet \vec{W} = \sum_i \underline{(\sum_{j=0}^{bw_x-1} \sum_{k=0}^{bw_w-1}
	2^{j+k} \times x_i[j] \times w_i[k])}$}
\label{eq:dot-product2}
\end{equation}
Conventional architectures rely on compute units that operate on all bits of individual operands, and require complex left-shift operations followed by wide-bitwidth additions as shown with an underline in Equation~\ref{eq:dot-product2}.
By leveraging the associativity property of the multiplication and addition, we can cluster the bit-wise operations that share the same significance position together and factor out the power of two multiplications.
In other words, this clustering can be realized by swapping the order of $\sum_i$ and $\sum_j \sum_k$ operators.
\begin{equation}
	\resizebox{0.75\columnwidth}{!}{
	$\vec{X} \bullet \vec{W} = \sum_{j=0}^{bw_x-1} \sum_{k=0}^{bw_w-1} 2^{j+k} \times
	\underline{(\sum_i x_i[j] \times w_i[k])}$}
\label{eq:dot-product-optimized}
\end{equation}
Leveraging this insight enables the use of significantly less complex, narrow-bitwidth, compute units (1-bit in the equation), exploiting bit-level parallelism, amortizing the cost of left-shift and wide-bitwidth addition.
Breaking down the dot-product is not limited to single bit and elements of the vectors can be \emph{bit-sliced} with different sizes.
%
As such, Equation~\ref{eq:dot-product-optimized} can be further generalized as:
\begin{equation}
	\resizebox{0.98\columnwidth}{!}{
    $ = \sum_{j=0}^{\frac{bw_x}{\alpha} -1} \sum_{k=0}^{\frac{bw_w}{\beta} -1} 2^{\alpha j+\beta k} \times$ \\ $\underline{(\sum_i
	 x_i[\alpha j:(\alpha+1)j] \times w_i[\beta k:(\beta+1)k])}$}
\label{eq:dot-product-optimized-generatlized}
\end{equation}
Here, $\alpha$ and $\beta$ are the bit-slices for operands $x_i$ and $w_i$ for the narrow-bitwidth compute units, respectively.
%
%
%
%
Figure~\ref{fig:compute_model}-(a) graphically illustrates bit-parallel vector composability using an example of a vector dot-product operation~($\vec{X} \bullet \vec{W} = \sum_{i} x_i \times w_i$) between two vectors of $\vec{X}$ and $\vec{W}$, each of which constitutes two 4-bit elements.
As it is shown with different shades, each element can be bit-sliced and broken down into two 2-bit slices.
With this bit-slicing scheme, the original element~($x_i$ or $w_i$) can be written as $2^2 \times bsl_{MSB} + 2^0 \times bsl_{LSB}$, where $bsl$s are the bit-slices and they are multiplied by powers of two based on their significance position.
With the aforementioned bit-slicing scheme, performing a product operation between an element from $\vec{X}$ and another from $\vec{W}$ requires four multiplications between the slices, each of which is also multiplied by the corresponding significance position factor.
However, because of the associativity property of the multiply-add, we cluster the bit-sliced multiplications that share the same significance position and apply the power of two multiplicands by shifting the accumulated result of the bit-sliced multiplications.

\niparagraph{Flexible-Bitwidth vector composability.}
%
%
The bit-parallel vector composable design style can enable \emph{flexible-bitwidth support at runtime}.
Figure~\ref{fig:compute_model}-(b) shows an example of flexible-bitwidth vector dot-product operation considering \emph{the same number of compute resources (2-bit multipliers, adders, and shifters)} as Figure~\ref{fig:compute_model}.
In Figure~\ref{fig:compute_model}-(b), a vector dot-product operation between a vector of inputs~($\vec{X}$) that has four 4-bit elements and a vector of weights~($\vec{W}$) that has four 2-bit elements is illustrated.
Using the same bit-slicing scheme, the original vector $\vec{X}$ is broken down to two sub-vectors that are required to go under dot-product with $\vec{W}$ and then get shifted and aggregated.
However, exploiting 2-bit datatypes for weights compared to 4-bit in the example given in Figure~\ref{fig:compute_model}-(a), provides $2\times$ boost in compute performance.
Bit-parallel vector composability enables maximum utilization of the compute resources, resulting in computing a vector dot-product operation with $2\times$ more elements using the same amount of resources.
%

The next section discusses the acceleration using this design style.

%

\section{Architecture Design for Bit-Parallel Vector Composability}
\label{sec:digital-composability}
%
%
To enable hardware realization for the insight of bit-parallel vector composability, the main building block of our design becomes a Composable Vector Unit (\cve), which performs the vector dot-product operation by splitting it into multiple \emph{narrow-bitwidth} dot-products.
As such, the \cvu consists of several Narrow-Bitwidth Vector Engines~(\lbvu) that calculate the dot-product of bit-sliced sub-vectors from the original vectors.
The \cvu then combines the results from \lbvus according to the bitwidth of each DNN layer.
Below, we discuss the micro-architecture.

%

%
\begin{figure}
	\centering
	\includegraphics[width=0.8\linewidth]{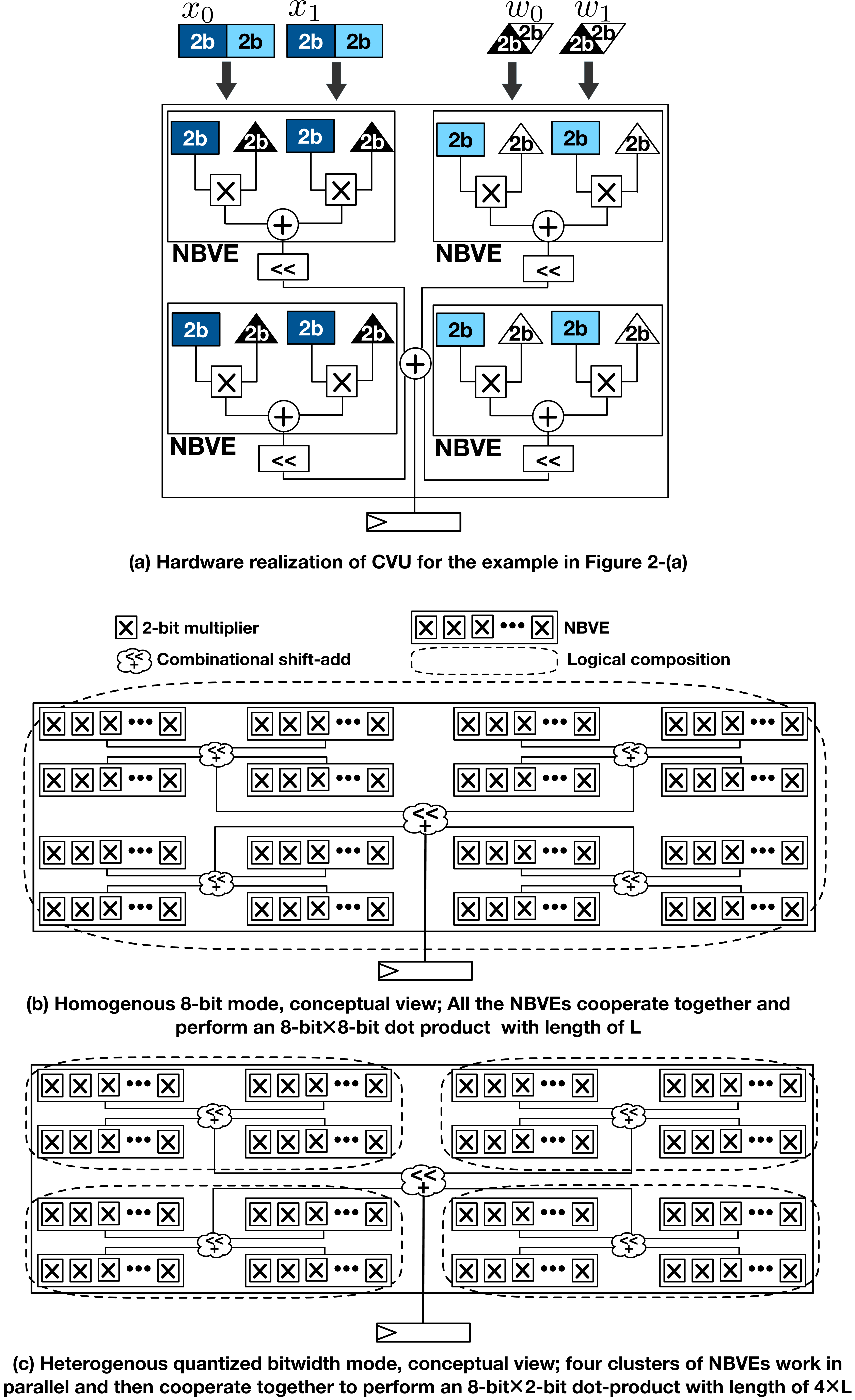} 
	\vspace{-1ex}
	\caption{Composable Vector Unit.}
 	\vspace{-4ex}
	\label{fig:digital-ve}
\end{figure}
\begin{figure*}[h]
\begin{minipage}{1\linewidth}
	\centering
	\includegraphics[width=1\linewidth]{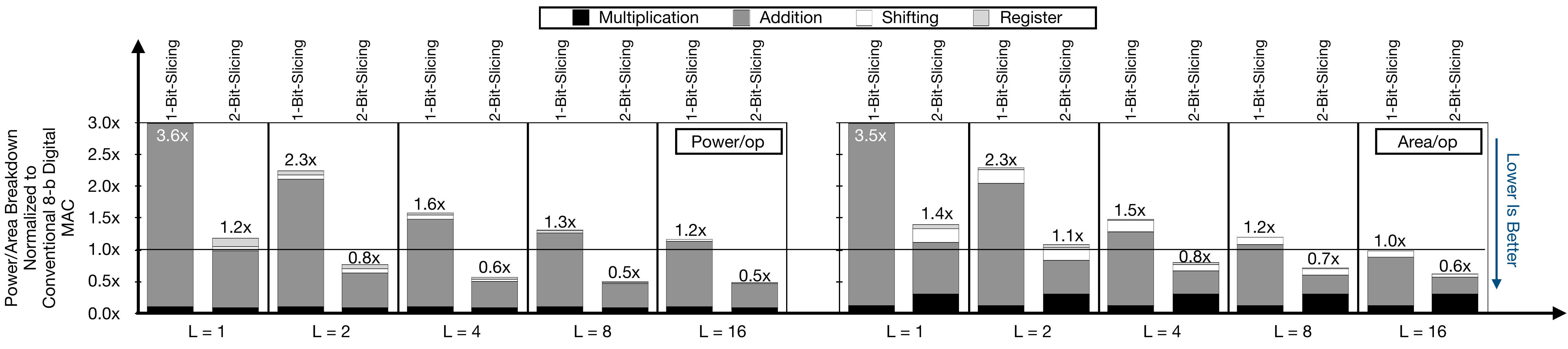} 
	\vspace{-3.5ex}
  \caption{Design space exploration for size of bit-slicing and vector lengths for vector composability.}
	\label{fig:dse-op-digital}
\end{minipage}
\vspace{-3ex}
\end{figure*}

\subsection{Composable Vector Unit (CVU)}
\label{sec:digital-cvu}
Figure~\ref{fig:digital-ve}-(a) illustrates an instance of the hardware realization of \cve for the vector dot-product example given in Figure~\ref{fig:compute_model}-(a).
%
%
%
In this example, \cve encapsulates \textsf{4} Narrow-Bitwidth Vector Engines~(\lbvu).
The number of \lbvus inside a \cve is based on the size of bit-slicing and maximum bitwidth of datatypes~(inputs and weights) that have been selected as two and four in this example, respectively.
A spatial array of narrow-bitwidth multipliers,  connected through an adder-tree, constitute each \lbvu.
Each narrow-bitwidth multiplier performs a 2-bit$\times$2-bit multiplication with a 2-bit-slice of inputs and weights, the result of which then goes to the adder-tree to get aggregated with the outputs of other multipliers inside an \lbvu.
As such, the \lbvu generates a single scalar that is the result of the dot-product operation between two bit-sliced sub-vectors.
The \cve then shifts the outputs of \lbvus based on the significance position of their bit-sliced operands and aggregates the shifted values across all the \lbvus to generate the final result of the vector dot-product operation.
In our design, we consider 8-bit as the maximum bitwidth of inputs and weights, commensurate with prior work~\cite{tpu:isca:2017}, and 2-bit slicing for the \cve.
As such, \cve encapsulates \textsf{16} \lbvus that work in parallel.
Below, we discuss how the \cvu operates when homogenous and heterogenous bitwidths are exploited.

\niparagraph{Homogeneous 8-bit mode of operation.}
Figure~\ref{fig:digital-ve}(b) illustrates the conceptual view of \cve when 8-bit datatypes are homogeneously exploited for DNN layers.
Each \lbvu performs a bit-parallel dot-product on 2-bit-sliced sub-vectors of the original 8-bit$\times$8-bit dot-product between vectors of length $L$, generating a scalar.
The \nbves are equipped with shifters to shift their scalar outputs.
Finally, to generate the final scalar of the 8-bit dot-product, all the \lbvus in a \cve,  globally cooperate together and \cve aggregates their outputs.

\niparagraph{Heterogeneous quantized bitwidth mode of operation.}
When the DNN uses heterogeneous bitwidths (less than 8-bit datatypes) across its layers, the \cvu can be dynamically reconfigured to match the bitwidths of the DNN layers at runtime. 
This includes both the reconfigurations of the shifters and the \lbvus composition scheme.
For instance, Figure~\ref{fig:digital-ve} (c) illustrates a case when 8-bit inputs and 2-bit weights are used, the \textsf{16} \lbvus will be clustered as four groups, each of which encapsulates four \lbvus.
In this mode, the \cve composes the \lbvus in two levels.
At the first level, all the four \lbvus in each cluster are privately composed together by applying the shift-add logic to complete a dot-product operation with a length of $L$.
At the second level, the \cve globally aggregates the outputs of all  four clusters and produces the scalar result of dot-product operation between vectors of length $4\times L$. 
As another example, if 2-bit datatypes are used for both inputs and weights, each \lbvu performs an independent dot-product, providing $16\times$ higher performance compared to the homogenous 8-bit mode.
%
%
To evaluate the benefits of the bit-parallel vector composability and its design tradeoffs, we perform a design space exploration for different number of multipliers in an \nbve (\textsf{L}) and choice of bit-slicing and analyze the sensitivity of the \cve's power/area to these parameters, as follows.
%
%

\subsection{Design Space Exploration and Tradeoffs}
Figure~\ref{fig:dse-op-digital} shows the design space exploration for 1-bit and 2-bit slicing, in addition to different lengths of vectors for \lbvu from $L=1$ to $L=16$.
All the design points in this analysis are synthesized in 500~Mhz and 45~nm node.
Y-Axis shows the power and area per one 8-bit $\times$ 8-bit MAC operation performed by \cve normalized to the power/area of a conventional digital 8-bit MAC unit.
We sweep the $L$ parameter for 1-bit slicing and 2-bit slicing in the X-axis.
%
%
%
Figure~\ref{fig:dse-op-digital} also shows the breakdown of power and area across four different hardware logics; multiplication, addition, shifting, and registering.
Inspecting this analysis leads us to following key observations:

\emph{(1) Adder-tree consumes the most power/area and might bottleneck the efficiency.}
As we observe in both 1-bit and 2-bit slicing, across all the hardware components adder-tree ranks first in power/area consumption.
Bit-parallel vector composability imposes two levels of add-tree logic:
(a) An adder-tree private to each \lbvu that sums the results of narrow-bitwidth multiplications.
(b) A global adder-tree that aggregates the outputs of \lbvus to generate the final scalar result.
Hence, to gain power/area efficiency, the cost of add-tree requires to be minimized.

\emph{(2) Integrating more narrow bitwidth multipliers within \lbvus~(exploiting DLP within BLP) minimizes the cost of add-tree logic.}
Encapsulating a larger number of narrow-bitwidth multipliers in an \nbve leads to amortizing the cost of add-tree logic across a wider array of multipliers and yields power/area efficiency.
%
%
As it is shown in Figure~\ref{fig:dse-op-digital}, increasing $L$ from \textsf{1} to \textsf{16} improves the power/area by $\approx 3\times$ in 1-bit slicing and $\approx 2.5\times$ in 2-bit slicing.
However, this improvement in power/area gradually saturates.
As such, increasing $L$ beyond 16 does not provide further significant benefits.

\emph{(3) The 2-bit slicing strikes a better balance between the complexity of the narrow-bitwidth multipliers and the cost of aggregation and operand delivery in \cvus.}
1-bit slicing requires 1-bit multipliers~(merely AND gates), that also generates 1-bit values as the inputs of the adder-trees in \lbvus.
However, slicing 8-bit operands to 1-bit require $8 \times 8 = 64$ \lbvus for a \cve, imposing a costly 64-input global adder-tree to \cves.
Consequently, as it is shown in Figure~\ref{fig:dse-op-digital}, 1-bit slicing does not provide any benefits compared to the conventional design. 
On the other hand, although 2-bit slicing results in generating 4-bit values by the multipliers as inputs to adder-trees in \lbvus, it quadratically decreases the total number of \lbvus in a \cve from 64 to 16.
This quadratic decrease trumps using wider-bitwidth values as the inputs of adder-trees in \nbves, significantly lowering add-tree cost.
In conclusion, the optimal design choice for a digital \cve comes with 2-bit slicing and length of $L=16$.
Compared to a conventional digital design, this design point provides $2.0\times$ and $1.7\times$ improvement in power and area respectively, for an 8-bit $\times$ 8-bit MAC operation.
Note that 4-bit slicing provides lower power/area for \cve design, however, it leads to underutilization of compute resources when DNNs with less than 4-bits are being processed.
As such, 2-bit slicing strikes a better balance between the efficiency of \cves and their overall utilization.

\emph{(4) Bit-parallel vector composability amortizes the cost of flexibility across the elements of vectors.}
Prior bit-flexible works, both spatial (e.g., BitFusion~\cite{bitfusion:isca18}) and temporal (e.g., Stripes and Loom~\cite{stripes:micro:2016, loom:arxiv:2017}), enable supporting deep quantized DNNs with heterogenous bitwidths at the cost of extra area overheads.
BitFusion~\cite{bitfusion:isca18} exploits spatial bit-parallel composability for a scalar.
This design can be assumed as one possible configuration of bit-parallel vector composability with 2-bit slicing and $L=1$.
As shown in Figure~\ref{fig:dse-op-digital}, this design point imposes 40\% area overhead as compared to conventional design, while our design provides 40\% reduction in area.
Also our proposed \cvu provides $2.4\times$ improvement in power as compared to Fusion Units in \bitfusion.
This result seems counter intuitive at the first glance as flexibility often comes with an overhead.
In fact, the cost of bit-level flexibility stems from aggregation logic that puts the results back together.
Our proposed bit-parallel vector-level composability amortizes this cost across the elements of the vector.
Moreover, since it reduces the complexity of the cooperating narrower bitwidth units, it leads to even further reduction.

\subsection{Overall Architecture}
Conceptually, bit-parallel vector composability is orthogonal to the architectural organization of the \cvus.
We explore this design style using a 2D systolic array architecture, which is efficient for matrix-multiplications and convolutions, as explored by prior works~\cite{tpu:isca:2017}.
In this architecture, called \bpvec, each \cvu reads a vector of weights from its private scratchpad, while a vector of inputs is shared across columns of \cvus in each row of the 2D array.
The scalar outputs of \cvus aggregate across columns of the array in a systolic fashion and accumulate using 64-bit registers.

\section{Evaluation}
\label{sec:eval}
\subsection{Methodology}
\label{sec:method}
\niparagraph{Workloads.}
\begin{table}
	\centering
	\caption{\\ \scshape Evaluated DNN models.}
	\vspace{-1.5ex}
	\includegraphics[width=1\linewidth]{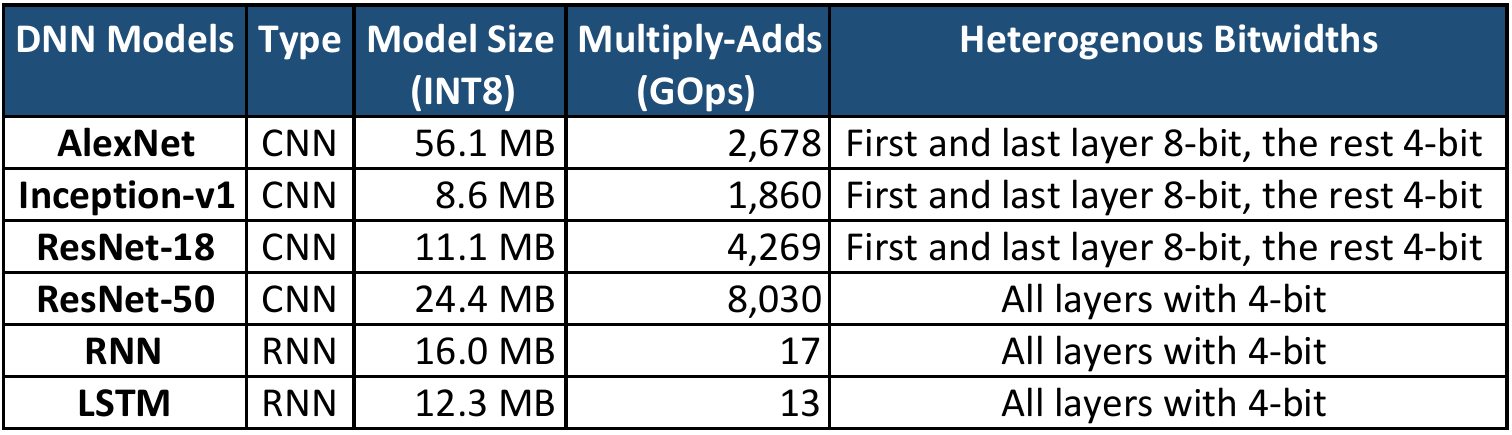} 
	\label{tab:benchmarks}
\vspace{-4.5ex}
\end{table}
%
%
Table~\ref{tab:benchmarks} details the specification of the evaluated models.
We evaluate our proposal using these neural models in two cases of homogenous and heterogenous bitwidths.
For the former one we use 8-bit datatypes for all the activations and weights and for the later we use the bitwidths reported in the results of the literature~\cite{wrpn, qnn:arxiv:2016, choi2018pact} that maintains the full-precision accuracy of the models.
%

\niparagraph{ASIC baselines.}
\begin{table}
	\centering
	\caption{\\ \scshape Evaluated hardware platforms.}
	\vspace{-1.5ex}
	\includegraphics[width=1\linewidth]{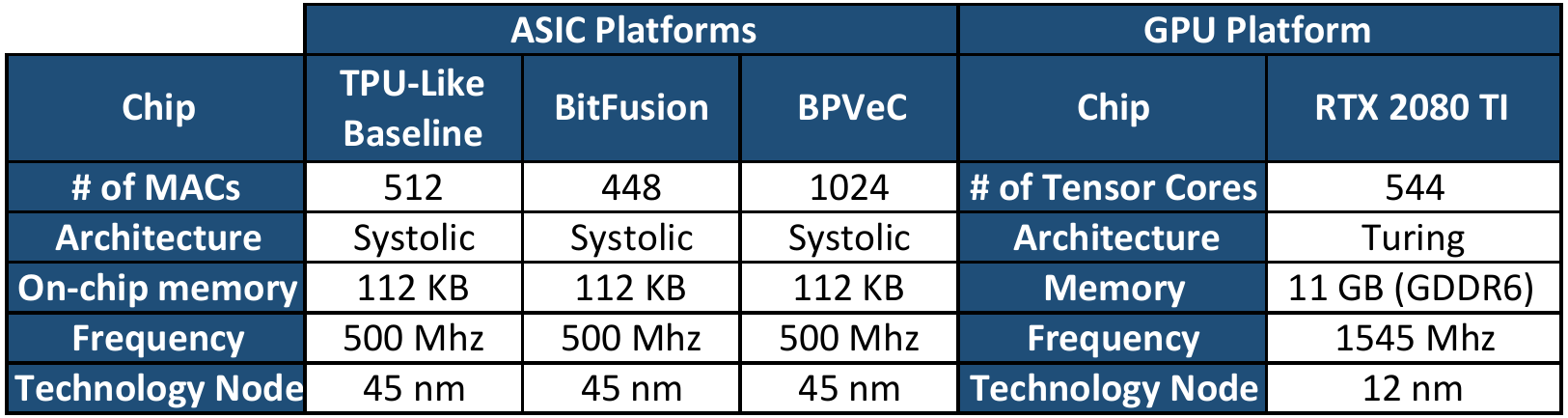} 
	\label{tab:platforms}
	\vspace{-6ex}
\end{table}
%
%
For the experiments with homogeneous fixed bitwidths, we use a TPU-like accelerator with a systolic architecture.
For the case of heterogeneous bitwidths, we use BitFusion~\cite{bitfusion:isca18}, a state-of-the-art spatial bit-flexible DNN accelerator, as the comparison point.
In all setups, we use 250~mW core power budget for all the baselines and the proposed accelerator in 45~nm technology node and with 500~Mhz frequency. 
Table~\ref{tab:platforms} details the specifications of the evaluated platforms.
We modify the open-source simulation infrastructure in~\cite{bitfusion:isca18} to obtain end-to-end performance and energy metrics for the TPU-like baseline accelerator, baseline \bitfusion accelerator, as well as the proposed \bpvec accelerator.
%

\niparagraph{GPU baseline.}
We also compare \bpvec to the Nvidia's RTX~2080~TI GPU, equipped with tensor cores that are specialized for deep learning inference.
Table~\ref{tab:platforms} shows the architectural parameters of this GPU.
For the sake of fairness, we use 8-bit execution for the case of homogenous bitwidths and 4-bit execution for heterogenous bitwidths using the Nvidia's \code{TensorRT 5.1} compiled with \code{CUDA 10.1} and \code{cuDNN 7.5}.
%
%
%

\niparagraph{Hardware measurements.}
We implement the proposed accelerator using Verilog RTL.
We use \code{Synopsis Design Compiler (L-2016.03-SP5)} for synthesis and measuring energy/area.
All the synthesis for the design space exploration presented in Figure~\ref{fig:dse-op-digital} are performed in 45~nm technology node and 500~Mhz frequency and all the design points meet the frequency criteria.
%
%
The on-chip scratchpads for ASIC designs are modeled with \code{CACTI-P}~\cite{cactip}.

\niparagraph{Off-chip memory.}
%
We evaluate our design style with both a moderate and high bandwidth off-chip memory system to assess its sensitivity to the off-chip bandwidth.
For moderate bandwidth, we use \code{DDR4} with 16 GB/sec bandwidth and 15~pJ/bit energy for data accesses.
We model the high bandwidth memory based on \code{HBM2} with 256 GB/sec bandwidth and 1.2 pJ/bit energy for data accesses~\cite{fine}.

%
%

\subsection{Experimental Results}
\label{sec:results}

\subsubsection{Without Bitwidth Heterogeneity}
%
%
\begin{figure}
	\centering
	\includegraphics[width=0.9\linewidth]{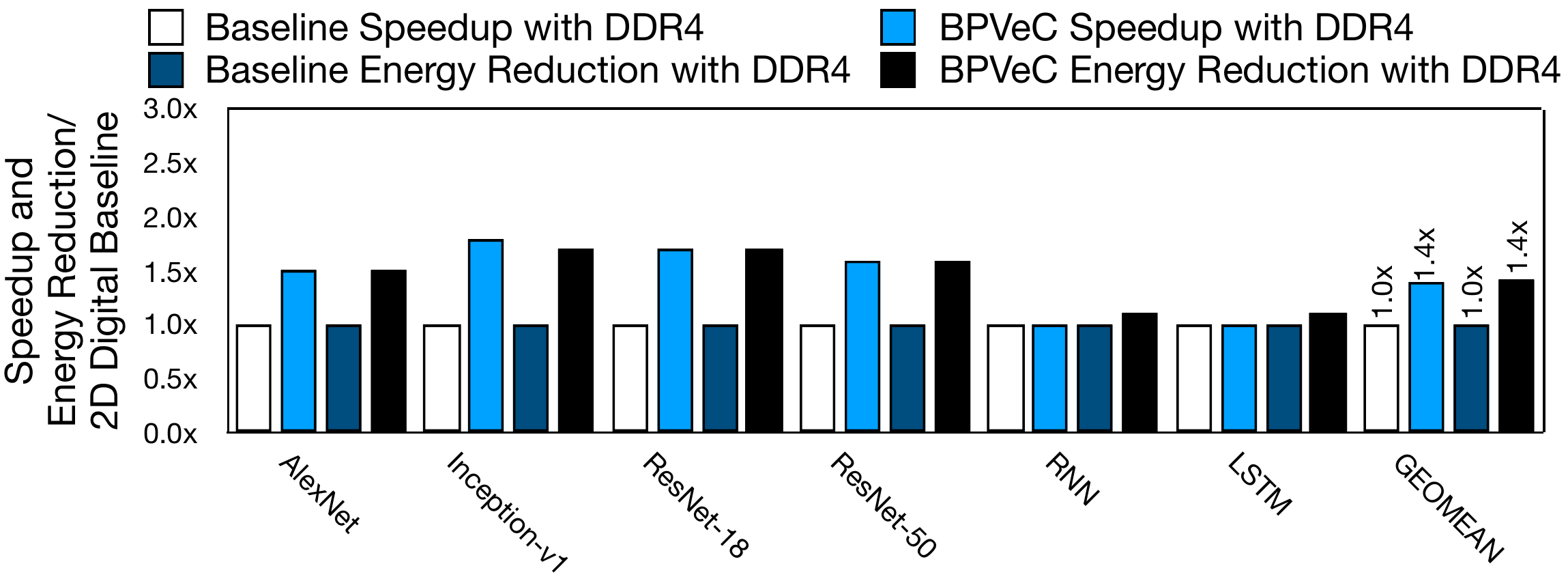} 
	\vspace{-1ex}
	\caption{Comparison to baseline; DDR4 memory and without bitwidth heterogeneity.}
	\label{fig:fixed-comp-2d}
\vspace{-3ex}
\end{figure}
\begin{figure}
	\centering
	\includegraphics[width=0.9\linewidth]{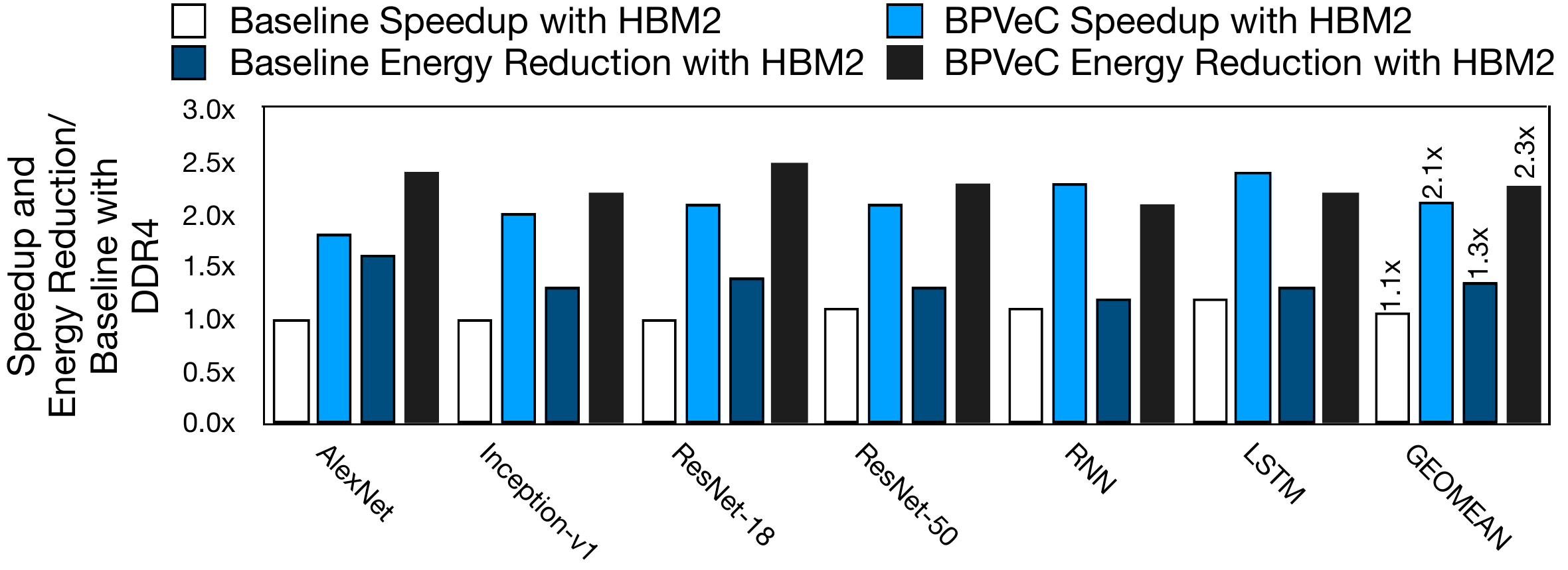} 
	\vspace{-1ex}
	\caption{Comparison to baseline; HBM2 memory and without bitwidth heterogeneity.}
	\label{fig:fixed-comp-3d}
\vspace{-4.5ex}
\end{figure}
%

Figure~\ref{fig:fixed-comp-2d} evaluates the performance and energy benefits of \bpvec across a range of DNNs with homogenous btiwidths~(8-bit).
The baseline uses the same systolic-array architecture as the \bpvec accelerator, but with conventional compute units that operate on individual operands and not bit-slices.
Both designs use a DRR4 memory system.
Bit-parallel vector composability enables our accelerator to integrate $\approx 2.0\times$ more compute resources compared to the baseline design under the same core power budget.
On average, \bpvec provides $40\%$ speedup and energy reduction.
%
%
Across the evaluated workloads, CNN models enjoy more benefits compared to RNN ones.
%
Unlike the CNNs that have significant data-reuse, the vector-matrix multiplications in RNNs have limited data-reuse and require extensive off-chip data accesses.
As such, the limited bandwidth of the DDR4 memory leads to starvation of the copious on-chip compute resources in \bpvec	.

Figure~\ref{fig:fixed-comp-3d} compares benefits from a high bandwidth memory (HBM2) for the baseline and \bpvec, normalized to the baseline with DDR4.
While benefits from high bandwidth memory are limited for the baseline design, our \bpvec enjoys a \speedupDigFixedThreeD and \energyDigFixedThreeD speedup and energy reduction, respectively.
Results suggest that \bpvec is better able to exploit the increased bandwidth and reduced energy cost from HBM2 memory system.
%
%
While all benchmarks see improved efficiency, benefits are highest for bandwidth-hungry \bench{RNN} and \bench{LSTM}.

\begin{figure}
	\centering
	\includegraphics[width=0.9\linewidth]{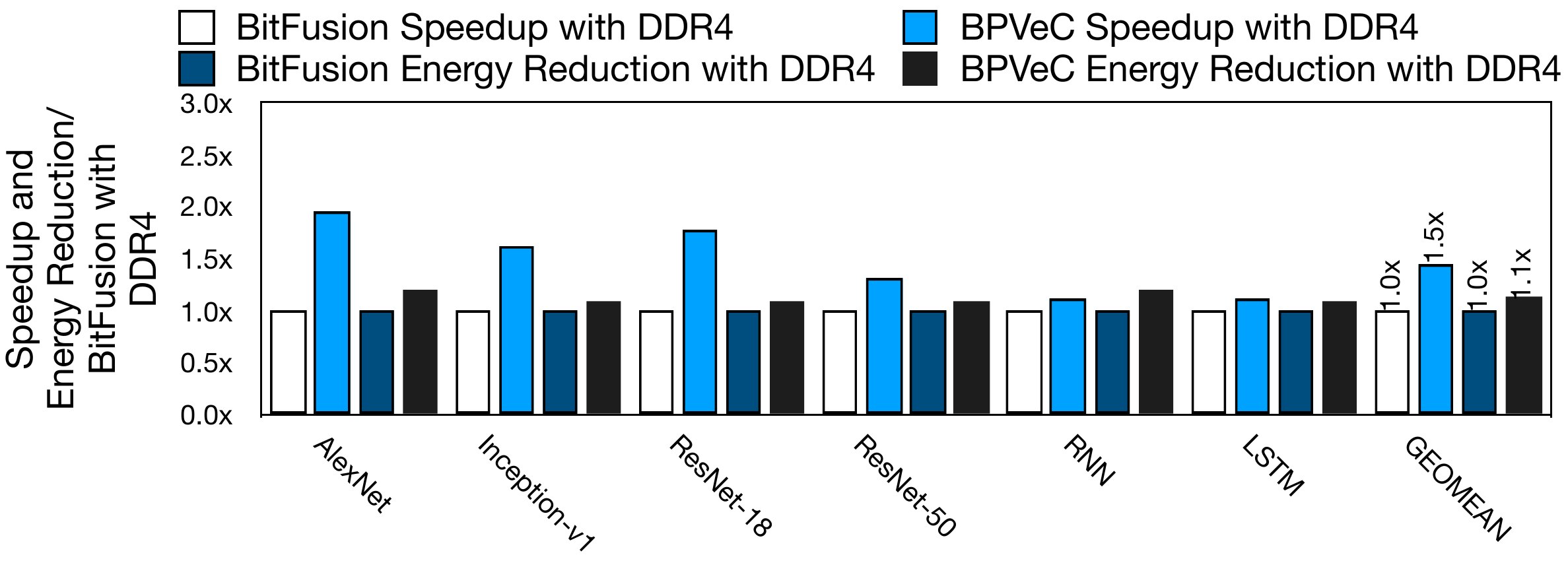} 
	\vspace{-1ex}
	\caption{Comparison to BitFusion; DDR4 memory and with bitwidth heterogeneity.}
	\label{fig:flexi-comp-2d}
	\vspace{-3ex}
\end{figure}

\subsubsection{With Bitwidth Heterogeneity}
\label{subsec:flexi}
Figure~\ref{fig:flexi-comp-2d} evaluates the performance and energy benefits for quantized DNNs with heterogenous bitwidths.
%
%
The baseline in Figure~\ref{fig:flexi-comp-2d} is \bitfusion~\cite{bitfusion:isca18}, a  state-of-the-art accelerator that also supports flexible bitwidths, but at the scalar level.
In this experiment, the baseline \bitfusion and \bpvec use DDR4 memory.
Bit-parallel vector composability enables our design to integrate $\approx 2.3\times$ more compute resources compared to \bitfusion under the same core power budget.
%
%
On average, \bpvec provides $50\%$ speedup and $10\%$ energy reduction over \bitfusion.
Across the evaluated workloads, CNN models enjoy more benefits compared to bandwidth-hungry \bench{RNN} and \bench{LSTM}.
%

\begin{figure}
	\centering
	\includegraphics[width=0.9\linewidth]{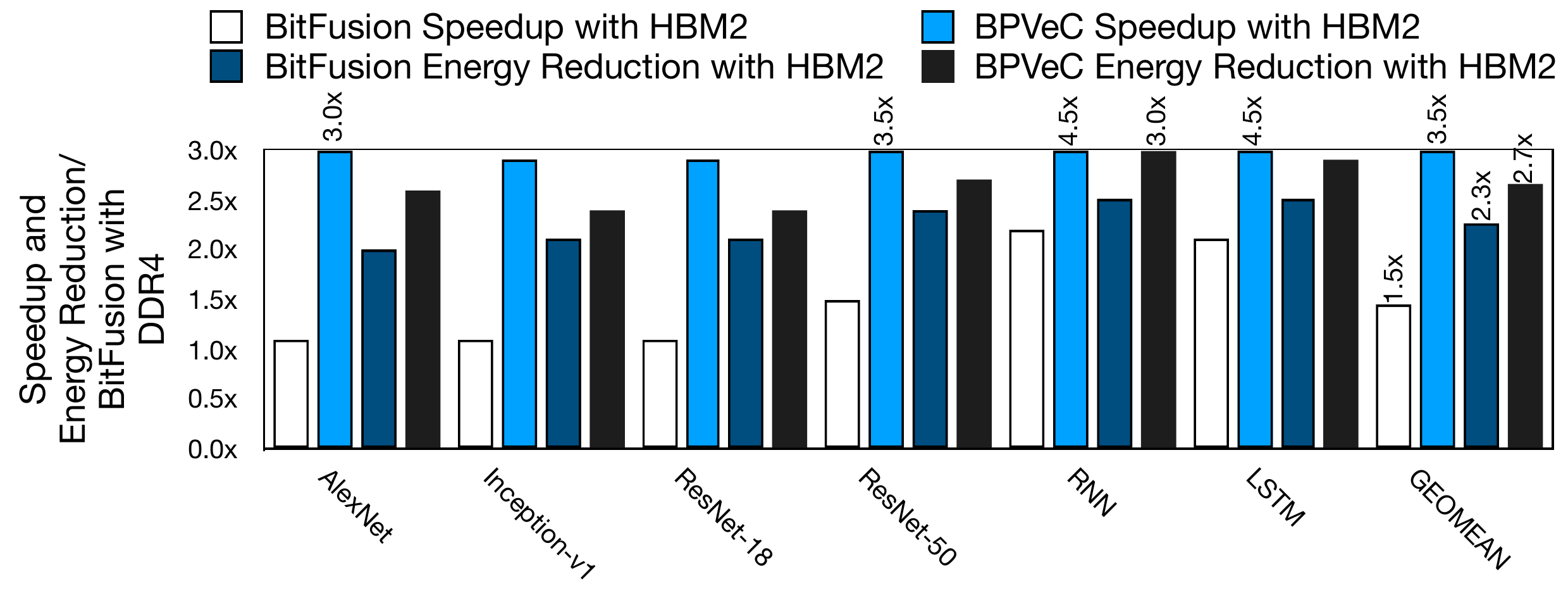} 
	\vspace{-1ex}
	\caption{Comparison to BitFusion; HBM2 memory and with bitwidth heterogeneity.}
	\label{fig:flexi-comp-3d}
\vspace{-4.5ex}
\end{figure}
%
Figure~\ref{fig:flexi-comp-3d} studies the interplay of high off-chip bandwidth with flexible-bitwidth acceleration.
The speedup and energy reduction numbers are normalized to \bitfusion with moderate bandwidth DDR4.
\bpvec provides \speedupFlexiDigThreeD speedup and $20\%$ energy reduction over  \bitfusion with HBM2 memory~($3.5\times$ speedup and $2.7\times$ energy reduction over the baseline 2D \bitfusion).
%
%
\bench{RNN} and \bench{LSTM}, see the highest performance benefits ($4.5\times$), since these benchmarks can take advantage of both the increased compute units in \bpvec design, as well as the increased bandwidth form HBM2.
%
%
%

\begin{figure}[h]
	\centering
	\includegraphics[width=0.85\linewidth]{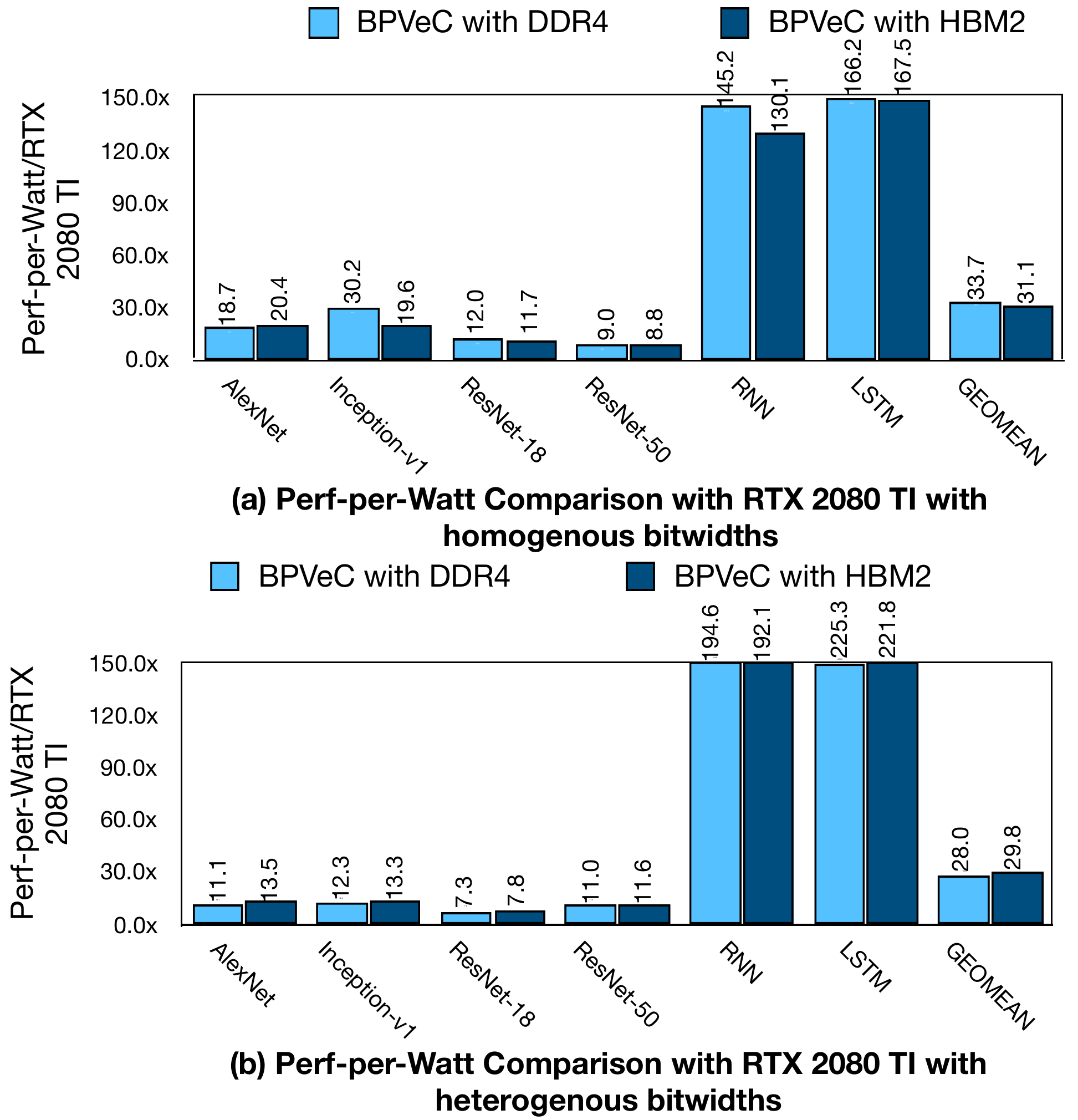} 	
	\vspace{-1.5ex}
	\caption{Performance-Per-Watt comparison to RTX 2080 TI GPU.}
	\label{fig:gpu}
\vspace{-4ex}
\end{figure}

\subsubsection{GPU Comparison}
%
%
Figure~\ref{fig:gpu} compares the Performance-per-Watt of \bpvec design with DDR4 and HBM2 memory and with homogenous (Figure~\ref{fig:gpu} (a)) and heterogenous (Figure~\ref{fig:gpu}) bitwidths for DNN layers, respectively as compared to the Nvidia's RTX~2080~TI GPU.
%
%
%
%
%
With homogenous bitwidths~(Figure~\ref{fig:gpu} (a)), \bpvec achieves $33.7 \times$ and $31.1 \times$ improvements on average with DDR4 and HBM2 memory, respectively.
Across the evaluated workloads, the RNN models see the most benefits.
These models require a large amount of vector-matrix multiplications, which particularly suitable for the proposed bit-parallel vector composability design style.
In the case of heterogenous bitwidths, the benefits go to $28.0 \times$ and $29.8 \times$ with DDR4 and HBM2, respectively.
%
%
The trends look similar to the homogenous 8-bit mode, since both the design points and the GPU baseline exploit the deep quantized mode of operations.
\section{Related Work}
\label{sec:related}
%
A large body of inspiring  work has explored hardware acceleration for DNNs by exploiting their algorithmic properties such as data-level parallelism, tolerance for reduced precision and sparsification, and redundancy in computations.
To realize the hardware accelerators, prior efforts have built upon isolated compute units that operate on all the bits of individual operands, and have used multiple compute units operating together to extract data-level parallelism.
This work introduces a different design style that explores the interleaving of bit-level operations across compute units in the context of \emph{vectors} and combine the benefits from bit-level parallelism and data-level parallelism, both of which are abundant in DNNs.
Below, we discuss the most related works.
%

\niparagraph{Design without support for bit-level composability.}
Prior works in TPU~\cite{tpu:isca:2017} and Eyeriss~\cite{eyeriss:isca:2016} design hardware accelerators to extract data-level parallelism in DNNs.
%
SCNN~\cite{scnn:isca:2017}, EIE~\cite{eie:isca:2016}, and Cnvlutin~\cite{cnvlutin:isca:2016} use both zero-skipping and data-level parallelism for efficient DNN execution.
%
%
%
Brainwave~\cite{brainwave:isac:2018} also uses SIMD vectorized execution to extract data-level parallelism on FPGAs, however with fixed-bitwidth execution whose bitwidth is decided before synthesizing the FPGA on the design.
ISAAC~\cite{isaac:isca:2016} and PRIME~\cite{prime:isca:2016} build upon ResistiveRam(ReRam) technology to provide high energy efficiency.
Further, ISAAC and PRIME operate on vectors of data in the analog (current) domain to mitigate the high cost of ADC.
In contrast, we focus on interleaving of the bit-level and data-level parallelism in vector units, in addition to hardware support for bitwidth heterogeneity.
%

\niparagraph{Design with support for bit-level flexibility through bit-serial computation.}
Stripes~\cite{stripes:micro:2016}, Loom~\cite{loom:arxiv:2017}, UNPU~\cite{unpu:isscc:2018} exploit the tolerance to reduced bitwidth in DNNs to yield performance benefits by exploring bit-\emph{bit serial compute units}.
The data-level parallelism compensates for bit-serial individual operations.
Our design in contrast interleaves bit-level parallelism with data-level parallelism.

\niparagraph{Designs with support for bit-level composability}
\bitfusion~\cite{bitfusion:isca18} uses bit-level parallelism with a spatial design. 
We provide a head-to-head comparison with \bitfusion in Section~\ref{subsec:flexi}.
Laconic~\cite{laconic:isca:2019} combines spatial bit-level composability with temporal execution to support for bit-sparsity and reduce the ineffectual computations.
These inspiring efforts do not focus on bit-parallel vector composability that breaks the calculations across spatial composable units that cooperate at the level of vectors.
%

\section{Conclusion}
\label{sec:conclusion}
\vspace{-1ex}
Traditionally, neural accelerators have relied on extracting DLP using isolated and self-sufficient compute units that process all the bits of operands.
This work introduced a different design style, \emph{bit-parallel vector-composability}, that operates on operand bit-slices to interleave and combine the traditional data-level parallelism with bit-level parallelism.
Across a range of deep models the results show that the proposed design style offers significant performance and efficiency compared to even bit-flexible  accelerators.
\section{Acknowledgement}
This work was in part supported by the National Science Foundation (NSF) awards CNS\#1703812, ECCS\#1609823, CCF\#1553192, Air Force Office of Scientific Research (AFOSR) Young Investigator Program (YIP) award \#FA9550-17-1-0274, National Institute of Health (NIH) award \#R01EB028350, and AirForce Research Laboratory (AFRL) and Defense Advanced Research Project Agency (DARPA) under agreement number \#FA8650-20-2-7009 and \#HR0011-18-C-0020. The U.S. Government is authorized to reproduce and distribute reprints for Governmental purposes not withstanding any copyright notation thereon. The views and conclusions contained herein are those of the authors and should not be interpreted as necessarily representing the official policies or endorsements, either expressed or implied of Google, Samsung, NSF, AFSOR, NIH, AFRL, DARPA or the U.S. Government.

\scriptsize

\bibliographystyle{IEEEtranS}
\bibliography{paper.bib}

\begin{thebibliography}{10}
\providecommand{\url}[1]{#1}
\csname url@samestyle\endcsname
\providecommand{\newblock}{\relax}
\providecommand{\bibinfo}[2]{#2}
\providecommand{\BIBentrySTDinterwordspacing}{\spaceskip=0pt\relax}
\providecommand{\BIBentryALTinterwordstretchfactor}{4}
\providecommand{\BIBentryALTinterwordspacing}{\spaceskip=\fontdimen2\font plus
\BIBentryALTinterwordstretchfactor\fontdimen3\font minus
  \fontdimen4\font\relax}
\providecommand{\BIBforeignlanguage}[2]{{%
\expandafter\ifx\csname l@#1\endcsname\relax
\typeout{** WARNING: IEEEtranS.bst: No hyphenation pattern has been}%
\typeout{** loaded for the language `#1'. Using the pattern for}%
\typeout{** the default language instead.}%
\else
\language=\csname l@#1\endcsname
\fi
#2}}
\providecommand{\BIBdecl}{\relax}
\BIBdecl
\renewcommand{\BIBentryALTinterwordstretchfactor}{4}

\bibitem{cnvlutin:isca:2016}
J.~Albericio \emph{et~al.}, ``Cnvlutin: ineffectual-neuron-free deep neural
  network computing,'' in \emph{ISCA}, 2016.

\bibitem{eyeriss:isca:2016}
Y.-H. Chen \emph{et~al.}, ``Eyeriss: A spatial architecture for
  energy-efficient dataflow for convolutional neural networks,'' in
  \emph{ISCA}, 2016.

\bibitem{prime:isca:2016}
P.~Chi \emph{et~al.}, ``Prime: A novel processing-in-memory architecture for
  neural network computation in reram-based main memory,'' in \emph{ISCA},
  2016.

\bibitem{choi2018pact}
J.~Choi \emph{et~al.}, ``Pact: Parameterized clipping activation for quantized
  neural networks,'' \emph{arXiv preprint arXiv:1805.06085}, 2018.

\bibitem{releq}
A.~T. Elthakeb \emph{et~al.}, ``Releq: an automatic reinforcement learning
  approach for deep quantization of neural networks,'' \emph{arXiv preprint
  arXiv:1811.01704}, 2018.

\bibitem{brainwave:isac:2018}
J.~Fowers \emph{et~al.}, ``A configurable cloud-scale dnn processor for
  real-time ai,'' in \emph{ISCA}, 2018.

\bibitem{eie:isca:2016}
S.~Han \emph{et~al.}, ``Eie: efficient inference engine on compressed deep
  neural network,'' in \emph{ISCA}, 2016.

\bibitem{qnn:arxiv:2016}
I.~Hubara \emph{et~al.}, ``Quantized neural networks: Training neural networks
  with low precision weights and activations,'' \emph{arXiv}, 2016.

\bibitem{tpu:isca:2017}
N.~P. Jouppi \emph{et~al.}, ``In-datacenter performance analysis of a tensor
  processing unit,'' in \emph{ISCA}, 2017.

\bibitem{stripes:micro:2016}
P.~Judd \emph{et~al.}, ``Stripes: Bit-serial deep neural network computing,''
  in \emph{MICRO}, 2016.

\bibitem{unpu:isscc:2018}
J.~Lee \emph{et~al.}, ``Unpu: A 50.6 tops/w unified deep neural network
  accelerator with 1b-to-16b fully-variable weight bit-precision,'' in
  \emph{ISSCC}, 2018.

\bibitem{cactip}
S.~Li \emph{et~al.}, ``{CACTI-P: Architecture-level Modeling for SRAM-based
  Structures with Advanced Leakage Reduction Techniques},'' in \emph{ICCAD},
  2011.

\bibitem{wrpn}
A.~K. Mishra \emph{et~al.}, ``{WRPN:} wide reduced-precision networks,''
  \emph{arXiv}, 2017.

\bibitem{fine}
M.~O’Connor \emph{et~al.}, ``Fine-grained dram: energy-efficient dram for
  extreme bandwidth systems,'' in \emph{MICRO}.\hskip 1em plus 0.5em minus
  0.4em\relax IEEE, 2017, pp. 41--54.

\bibitem{scnn:isca:2017}
A.~Parashar \emph{et~al.}, ``{SCNN: An Accelerator for Compressed-sparse
  Convolutional Neural Networks},'' in \emph{ISCA}, 2017.

\bibitem{codex}
M.~Samragh \emph{et~al.}, ``Codex: Bit-flexible encoding for streaming-based
  fpga acceleration of dnns,'' \emph{arXiv preprint arXiv:1901.05582}, 2019.

\bibitem{isaac:isca:2016}
A.~Shafiee \emph{et~al.}, ``Isaac: A convolutional neural network accelerator
  with in-situ analog arithmetic in crossbars,'' in \emph{ISCA}, 2016.

\bibitem{loom:arxiv:2017}
S.~Sharify \emph{et~al.}, ``Loom: Exploiting weight and activation precisions
  to accelerate convolutional neural networks,'' \emph{arXiv}, 2017.

\bibitem{laconic:isca:2019}
S.~Sharify \emph{et~al.}, ``Laconic deep learning inference acceleration,'' in
  \emph{ISCA}, 2019.

\bibitem{bitfusion:isca18}
H.~Sharma \emph{et~al.}, ``Bit fusion: Bit-level dynamically composable
  architecture for accelerating deep neural network,'' in \emph{ISCA}.\hskip
  1em plus 0.5em minus 0.4em\relax IEEE, 2018, pp. 764--775.

\end{thebibliography}




\end{document}